\documentclass{article}

\usepackage{arxiv}

\usepackage{cite}
\usepackage{graphicx}
\usepackage{placeins}
\usepackage[bookmarks=false]{hyperref}
\usepackage{verbatim}  
\usepackage{subfig}
\usepackage{amsmath,amsfonts,amssymb,amsthm}
\usepackage{mathtools}
\usepackage{braket}
\usepackage{algorithm,algpseudocode}
\usepackage{url}
\usepackage[utf8x]{inputenc}
\usepackage{lipsum}
\usepackage{mwe}
\usepackage{booktabs}
\usepackage[normalem]{ulem}
\usepackage{color}
\usepackage{subfig}

\usepackage[
  separate-uncertainty = true,
  multi-part-units = repeat
]{siunitx}

\def\BibTeX{{\rm B\kern-.05em{\sc i\kern-.025em b}\kern-.08em
    T\kern-.1667em\lower.7ex\hbox{E}\kern-.125emX}}

\graphicspath{ {./images/} }  

\useunder{\uline}{\ul}{}

\newcommand{\domain}{\mathcal{X}}
\newcommand{\acc}{\ensuremath{f_{\text{acc}}}}

\begin{document}
\title{Neural Network Design: Learning from Neural Architecture Search}

\author{
Bas van Stein \\
LIACS, Leiden University\\
b.van.stein@liacs.leidenuniv.nl
\And
Hao Wang\\
LIP6, Sorbonne Universit{\'e}\\
hao.wang@lip6.fr
\And
Thomas B{\"a}ck \\
LIACS, Leiden University\\
t.h.w.baeck@liacs.leidenuniv.nl
}


%
\maketitle              
\begin{abstract}
Neural Architecture Search (NAS) aims to optimize deep neural networks' architecture for better accuracy or smaller computational cost and has recently gained more research interests. 
Despite various successful approaches proposed to solve the NAS task, the landscape of it, along with its properties, are rarely investigated. In this paper, we argue for the necessity of studying the landscape property thereof and propose to use the so-called Exploratory Landscape Analysis (ELA) techniques for this goal. Taking a broad set of designs of the deep convolutional network, we conduct extensive experimentation to obtain their performance. Based on our analysis of the experimental results, we observed high similarities between well-performing architecture designs, which is then used to significantly narrow the search space to improve the efficiency of any NAS algorithm. Moreover, we extract the ELA features over the NAS landscapes on three common image classification data sets, MNIST, Fashion, and CIFAR-10, which shows that the NAS landscape can be distinguished for those three data sets. Also, when comparing to the ELA features of the well-known Black-Box Optimization Benchmarking (BBOB) problem set, we found out that the NAS landscapes surprisingly form a new problem class on its own, which can be separated from all $24$ BBOB problems. Given this interesting observation, we, therefore, state the importance of further investigation on selecting an efficient optimizer for the NAS landscape as well as the necessity of augmenting the current benchmark problem set.
\end{abstract}


\section{Introduction}
\label{sec:intro}

Deep learning and deep neural networks (DNNs) are becoming more and more popular due to the way they can automatically perform feature extraction from raw data \cite{rawat2017deep}. These deep neural networks have shown great performance in complex tasks such as image classification and audio analysis. Especially with the use of Convolutional Neural Networks (CNN) \cite{krizhevsky2012imagenet}, on which we will focus during this study.

However, for these networks to perform well, good network architectures and network training methods (optimization methods) are required. These complex and deep architectures are till recently, mostly developed by hand. Recent advances in automatic machine learning (AutoML) also enabled the automatic development of these architectures, called Neural Architecture Search (NAS) \cite{Elsken2019,van2019automatic,rusu2016progressive}.

Using NAS techniques it is possible to build a well performing network in a matter of days without any knowledge of the problem and neural network architecture design.
However, there are also quite a few draw-backs to the way NAS techniques are being used. First of all, it requires a lot of computational time and resources to perform NAS experiments \cite{schwartz2019green}. Secondly, most of the results from the architecture search are not used and the algorithms do not learn anything that can be applied on multiple problem instances.

A commonly applied solution when there is not much data or a lot of computational power to develop new architectures, is to use an existing pre-trained architecture and fine tune this network on the data of a specific problem. This solution is also called transfer learning, where the knowledge gained by training on one data set is transferred to a new data set in the hope that the network can apply the same features to the new data set.
Recently, in \cite{Lu2020}, Neural Architecture Transfer (NAT), was proposed where the authors apply transfer learning on architectures by using a so called supernet. The supernet can be incrementally modified and is used to select task specific subnets.
While NAT already solves some of the computational issues regarding the development of neural architectures within a specific class of problems, it still cannot be used for solving completely different problem classes efficiently.

One of the main reasons why Deep Neural Networks and transfer learning is actually working, is that the optimization problem (fitting the network to the data) is very robust. There are many network architectures that can solve the same problem. We even dare to claim that there are an infinite number of neural network architectures that can solve a specific image classification problem with a decent accuracy. This makes it possible to find well working architectures by hand, or automatically using a small number of evaluations.
It also shows us, that we can learn much more from these network architectures and probably design much smaller and more efficient networks.

In this paper we analyse the optimization landscapes of Neural Architecture Search for three well known image classification data sets. The aim of this research is to extract rules of thumb for designing neural architectures and to use the information to develop more efficient NAS or NAT algorithms.

\begin{figure*}[!t]
\centering
\includegraphics[width=\textwidth]{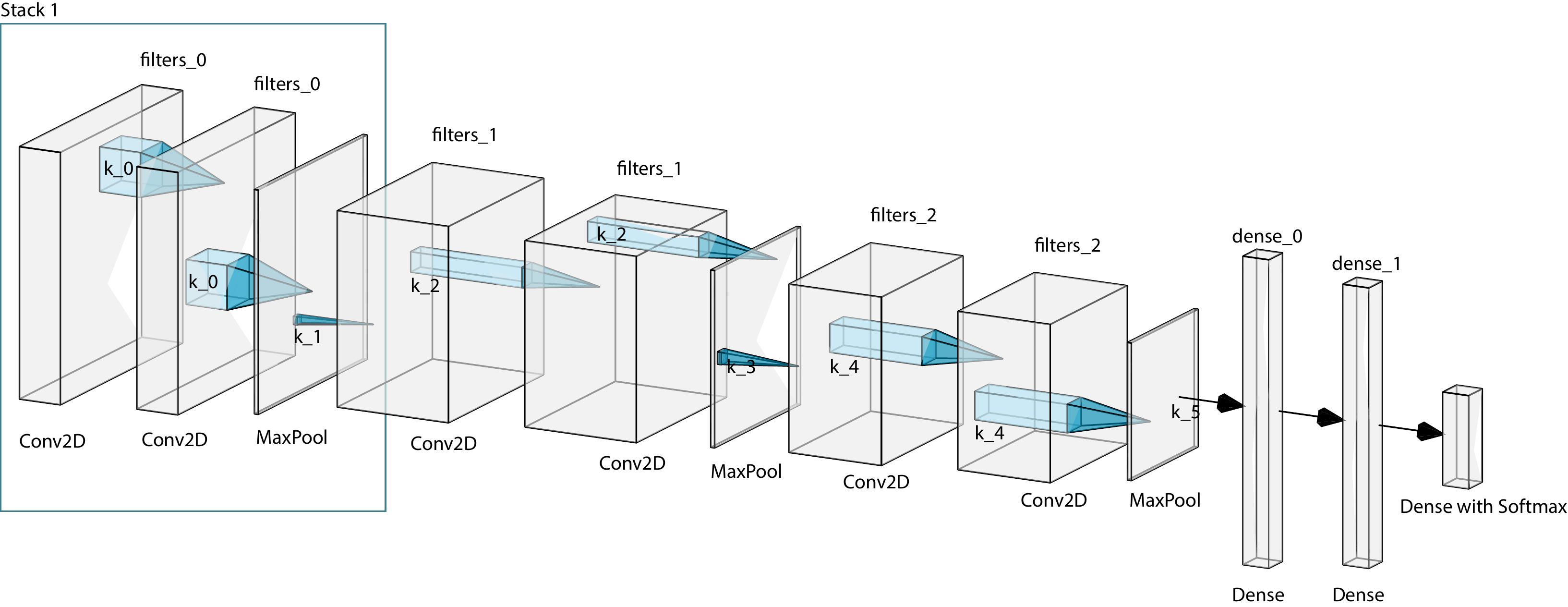}
\caption{The base architecture investigated in this paper: its feedforward structure starts with three stacks of layers, each of which comprises two convolutional layers and one max-pooling layer, and it ends with two layers of dense units and a softmax unit as the output. Note that, in each stack, the hyper-parameters such as the number of filters, strides, and the kernel size are independent from other stacks, and those two \texttt{Conv2D} layers always have the same number of filters.}\label{fig:base}
\end{figure*}

\section{Backgrounds and Problem Statement}
In the literature, various approaches have been proposed to handle the NAS problem, e.g., random search \cite{li2020random}, Bayesian optimization \cite{kandasamy2018neural}, evolutionary methods \cite{stanley2002evolving}, and reinforcement learning \cite{zoph2016neural}. 
Please see~\cite{ElskenMH19} for a comprehensive review of this topic. Despite the fruitful results in NAS, the landscape of NAS, along with its properties, are rarely investigated to the best of our knowledge. In this paper, we argue for the necessity of studying the landscape property of NAS, which potentially allows us to understand how hard such a search task practically is and where the difficulty lies in (e.g., the existence of discontinuity, saddle points, and non-convexity, etc.), and would pave the way towards the automated selection of optimizers for the NAS task.

\subsection{Explorative Landscape Analysis}
\label{subsec:ela}
For a real-valued target function, $f \colon \domain \xrightarrow{} \mathbb{R}$, Explorative Landscape Analysis (ELA)~\cite{MersmannBTPWR11,MunozKH15,DerbelLVAT19} aims to characterize certain features of this function, using some sample points in $\domain$ and the corresponding function values on $f$. For the so-called continuous optimization problem (i.e., $\domain \subseteq \mathbb{R}^d$), a large set of landscape features has already been devised to capture the information on dispersion, convexity, level sets, curvature, and ruggedness, which is extensively used in the automated algorithm selection~\cite{KerschkeT19}, to understand the search space~\cite{SkvorcEK20}, and for multi-objective optimization problems~\cite{KerschkeT16}. In this paper, we delve into the feature of performance indicators (e.g., testing accuracy) as the function of network architectures, by evaluating a huge number of randomly-generated network architectures and extracting the ELA features based on the performance of such architectures.

\subsection{Research Questions}
The problem that we are addressing in this paper is the time and cost inefficient approach of automatically designing and evolving well performing neural architectures. To improve upon the state-of-the-art approaches, such as NAT, and Meta-Neural Architecture Search \cite{wang2020m}, we have to gain additional knowledge of the underlying optimization landscape. 
\emph{``What are the characteristics of the objective space?"}. Characteristics such as the number of local optima, how are those optima located and how easily is it to get trapped in such optima, as well as smoothness, ruggedness etc.
\emph{``Are the objective space landscape characteristics shared between instances?"}. If some of the characteristics are shared between several problem instances, or even all NAS problems, this can be used to narrow down the search space or develop more generic approaches. 

Knowing the answers to these questions allows us to explore different optimization strategies and apply better fitted algorithms for architecture search and transfer.
Additional information about this landscape for a specific set of problems, such as image classification, allows us to establish ``rules of thumb" for a good starting neural architecture.

In this paper we focus mainly on three well-known image classification tasks, MNIST digits, Fashion and CIFAR-10. In the next section the empirical setup to analyse these NAS problem instances is explained in detail.

\section{Experimental Setup}
\label{sec:exp}

\paragraph{Design Space of Architectures} In this study, we narrowed down the design space to a feedforward convolutional network with stacks of convolutional layers, which is depicted in Fig.~\ref{fig:base}. Essentially, it entails three stacks of convolutional layers, followed by two layers of densely connected \textsc{relu} units and a \textsc{softmax} unit for the output. Each stack consists of two layers of convolutional filters (those two layers always have the same number of filters), one max-pooling layer, and one dropout layer to prevent over-fitting. We define the design space of architectures by varying the following hyper-parameters,
\begin{enumerate}
    \item for each convolutional layer, the number of filters, the size and stride of each filter kernel, the $L^2$ kernel regularization coefficient, and the dropout rate,
    \item for each densely connected layer, the number of units and the dropout rate, and
    \item the learning rate and the $L^2$ regularization coefficient.
\end{enumerate}
Each network is trained using the categorical crossentropy as loss function and using Stochastic Gradient Descent (SGD) with momentum as optimizer. The learning rate of SGD is one of the hyper-parameters as well.

We summarize those hyper-parameters and their sampling ranges in Table~\ref{tab:searchspaces}, where large sampling ranges are taken, allowing for generating architectures that are either very ``deep'' or ``shallow''. Taking the naming convention given in this table, we express the design space as: 
$$\mathcal{X} = \texttt{filters\_[0-2]}\times  \texttt{k\_[0-5]} \times \cdots \times \texttt{l2},$$
which is a \emph{$23$-dimensional space consisting of integers and real values}.
Moreover, we include two sets of different ranges for each parameter: The ``Initial Range'' is relatively wider, which is meant for exploring the NAS landscape with a good coverage and ``Reduced Range'' is narrowed down intentionally to focus on interesting regions of the design space, which is obtained by analysing the range of parameters of the top-50 architectures in the initial exploration ($5\%$ from 1000 in terms of the classification accuracy, please see below).

\begin{table}[!ht]
\centering
\setlength{\tabcolsep}{1.5pt}
\begin{tabular}{l | l | r | r | c }
\hline
Hyper-parameter & Naming & Initial Range & Reduced Range & Nr. \\
\hline
\#Filters & \texttt{filters\_[0-2]} & $(10, 600]$ & $(250, 400]$ & 3 \\
Kernel size & \texttt{k\_[0-5]} & $(1, 8]$ & $(3, 7]$ & 6\\
Strides & \texttt{s\_[0-2]} & $(1, 5]$ & $(2, 5]$ & 3 \\
\#Dense nodes & \texttt{dense\_size\_[0-1]} & $(0, 2000]$  & $(500, 1500]$ & 2\\
Dropout rate & \texttt{dropout\_[0-6]} & $(1\text{e-}5, 9\text{e-}1]$ & $(1\text{e-}1, 4\text{e-}1]$ & 7\\
Learning rate & \texttt{lr} & $(1\text{e-}5, 1\text{e-}2]$  & $(4\text{e-}3, 9\text{e-}3]$ & 1 \\
$L^2$ coeff. & \texttt{l2} & $(1\text{e-}5, 1\text{e-}2]$  & $(5\text{e-}4, 3\text{e-}3]$ & 1\\
\hline
\end{tabular}
\caption{\label{tab:searchspaces}The design spaces of network architectures for the initial experimentation (``Initial Range'') and the refined one (``Reduced Range''). Note that, for each layer, its hyper-parameter is independent from other layers and we denote by \emph{Nr.} the number of hyper-parameters of each kind. Also, each architecture uses \texttt{softmax} for the last activation function and \texttt{relu} for the convolutional layers.}%
\end{table}

\paragraph{Sample Size and Sampling Strategy}
Determining a proper sample size and the sampling strategy for the \emph{Design of Experiments} (DoEs) is still a daunting task in ELA~\cite{DBLP:journals/corr/abs-2006-11135,KerschkePWT16}. We decided to use a relatively large sample size of $1000$ (compared to the $23$-dimensional design space), which would yield reliable feature values (i.e., with acceptable dispersion), despite the enormous computational cost it incurred. Moreover, in~\cite{DBLP:journals/corr/abs-2006-11135}, it is shown that the landscape features are also susceptible to the sampling strategy and Latin Hypercube Sampling (LHS)~\cite{mckay2000comparison} is more statistically robust among the conventional sampling/design methods. Hence, we have chosen LHS as the sampling strategy in this study. Throughout this paper, we shall use $\acc, X\subseteq \mathcal{X}, y = \acc(X)$ to denote the performance indicator, the design, and the performance value of network architectures, respectively.





\paragraph{Training Data Sets and Computation Details}
Each configuration from the design of experiments is evaluated on three very popular image classification benchmarks, MNIST \cite{lecun1998gradient}, FASHION \cite{xiao2017fashion} and CIFAR-10 \cite{krizhevsky2009learning}\footnote{The DOE, source code, and the experimental results are available at \url{https://github.com/Basvanstein/LearningFromNAS} \cite{bas_van_stein_2020_4043005}.}.
We use a 20\% validation set to calculate the accuracy as objective metric. The networks are trained using the SGD optimizer on batches of $100$ samples using $50$ epochs.

\paragraph{ELA feature computation}
In the context of continuous optimization problems, more than $300$ numerical features have been proposed, which are implemented in the \verb|R|-package \textbf{flacco}\footnote{\url{https://github.com/kerschke/flacco}.}~\cite{KerschkeT16}.
In this paper, we manually select $20$ features out of $300$ for characterizing the landscape of NAS, which settle in five categories. Given the design $X$ of architectures and its performance values $y$, those selected features are defined as follows:
\begin{itemize}
    \item The \textbf{dispersion} of all the design points in $X$ and that of some top-ranked design points. In this study, we take the top-2\% and 5\% subset of $X$ (in terms of \acc).
        \begin{itemize}
            \item \texttt{disp.diff\_mean\_[02|05]}: the difference between the dispersion of $X$ and that of top-2\% or 5\% subset.
            \item \texttt{disp.ratio\_mean\_[02|05]}: the ratio between the dispersions described above.
        \end{itemize}

    It can be observed that the distances between the top $2$ and $5$ percent with respect to the distances between the complete DOE are close to each other (close to $1.0$), but instances in the top $2$ percent are slightly closer to each other. 
    \item The $y$-\textbf{distribution} features, \texttt{distr.kurtosis} and \texttt{distr.skewness} compute the kurtosis and skewness of $y$, respectively.
    \item The \textbf{Information Content of Fitness Sequences} (ICoFiS) features~\cite{6719480} quantify, through a random walk in $\mathcal{X}$, the smoothness, ruggedness, and neutrality of the landscape. For instance, when taking a random walk $(x_1,x_2,\ldots)$ in $X$, we can evaluate the maximal entropy (\texttt{ic.h.max}) of consecutive fluctuations of the corresponding $y$-value, i.e., $y_{i+1} - y_i, i=1,2,\ldots$ (Note that such a difference is discretized. Please see~\cite{MunozKH15} for details). All information content features are named as \texttt{ic.*} in the following discussions and we omit the description of those for brevity.
    \item The \textbf{meta-model} features take some important properties (e.g., adjusted $R^2$) of some simple meta-models trained on $(X,y)$.
        \begin{itemize}
            \item \texttt{lin\_simple.adj\_r2}: the adjusted $R^2$ of a simple linear model (w/o interactions between variables).
            \item \texttt{lin\_simple.intercept}: intercept of the linear model described above.
            \item \texttt{lin\_w\_interact.adj\_r2}: the adjusted $R^2$ of a simple linear model (w/ interactions between variables).
            \item \texttt{quad\_simple.adj\_r2}: the adjusted $R^2$ of a quadratic model.
        \end{itemize}
    \item The \textbf{nearest better clustering} features compare, for each $x\in X$, the distance from $x$ to its nearest neighbor and that to their \emph{nearest better neighbor}, that is the nearest neighbor having a better $y$-value than $x$.  All nearest better clustering features are named according to \texttt{nbc.*} in the following discussions. Please see~\cite{KerschkeT2019flacco} for the detailed description of them.
\end{itemize}


\section{Visualizing the NAS Landscape}
\label{sec:results}
In the first experiment, we generated a huge DOE representing $1000$ different network architectures, where each architecture is trained with $50$ epochs and the same (relatively simple) architecture of three stacks with in each stack two convolutional layers (Figure \ref{fig:base}).
\begin{figure*}[!t]
\centering
\subfloat[][MNIST + initial range]{\includegraphics[width=.33\textwidth,trim=0mm 0mm 0mm 7mm,clip]{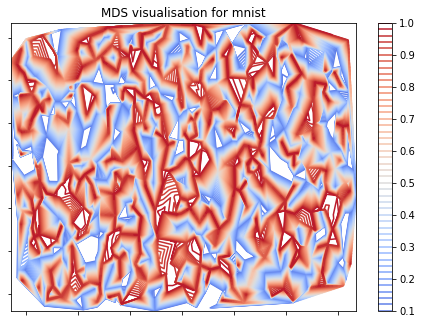}\label{fig:MDS-mnist1}}
\subfloat[][Fashion + initial range]{\includegraphics[width=.33\textwidth,trim=0mm 0mm 0mm 7mm,clip]{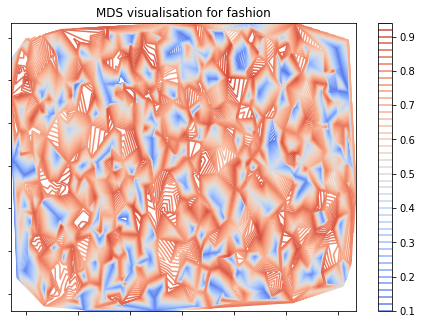}\label{fig:MDS-fashion1}}
\subfloat[][CIFAR-10 + initial range]{\includegraphics[width=.342\textwidth,trim=0mm 0mm 0mm 7mm,clip]{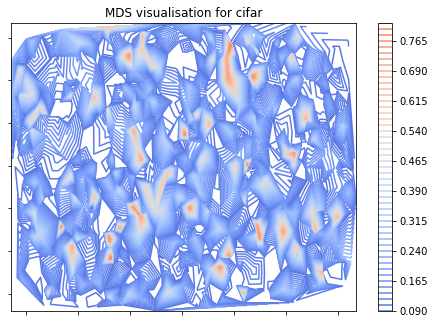}\label{fig:MDS-cifar1}}\\ \vspace{-3mm}
\subfloat[][MNIST + reduced range]{\includegraphics[width=.33\textwidth,trim=0mm 0mm 0mm 7mm,clip]{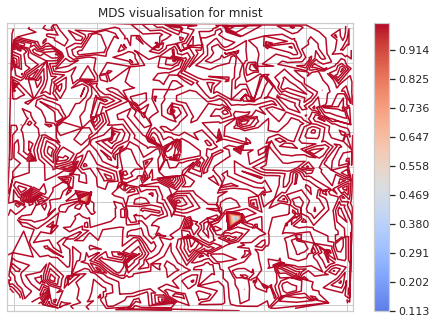}
\label{fig:MDS-mnist2}}
\subfloat[][Fashion + reduced range]{\includegraphics[width=.33\textwidth,trim=0mm 0mm 0mm 7mm,clip]{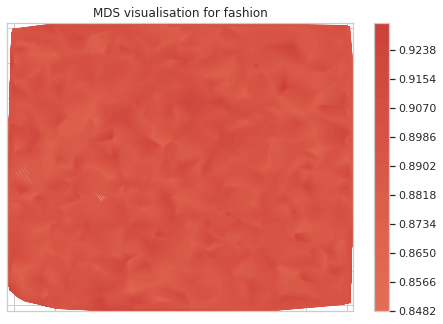}\label{fig:MDS-fashion2}}
\subfloat[][CIFAR-10 + reduced range]{\includegraphics[width=.33\textwidth,trim=0mm 0mm 0mm 7mm,clip]{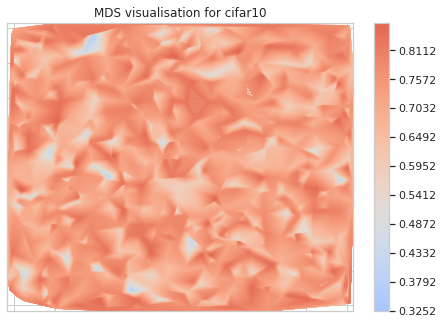}\label{fig:MDS-cifar2}}
\caption{Rendered by Multi Dimensional Scaling (MDS), the $2D$ landscape of classification accuracy as a function of network architectures, where the architectures are created using the initial range (top row) and the reduce range (bottom row) and the accuracy value is indicated by the color scale.}
\end{figure*}

\begin{figure*}[!t]
\centering
\includegraphics[width=.95\textwidth]{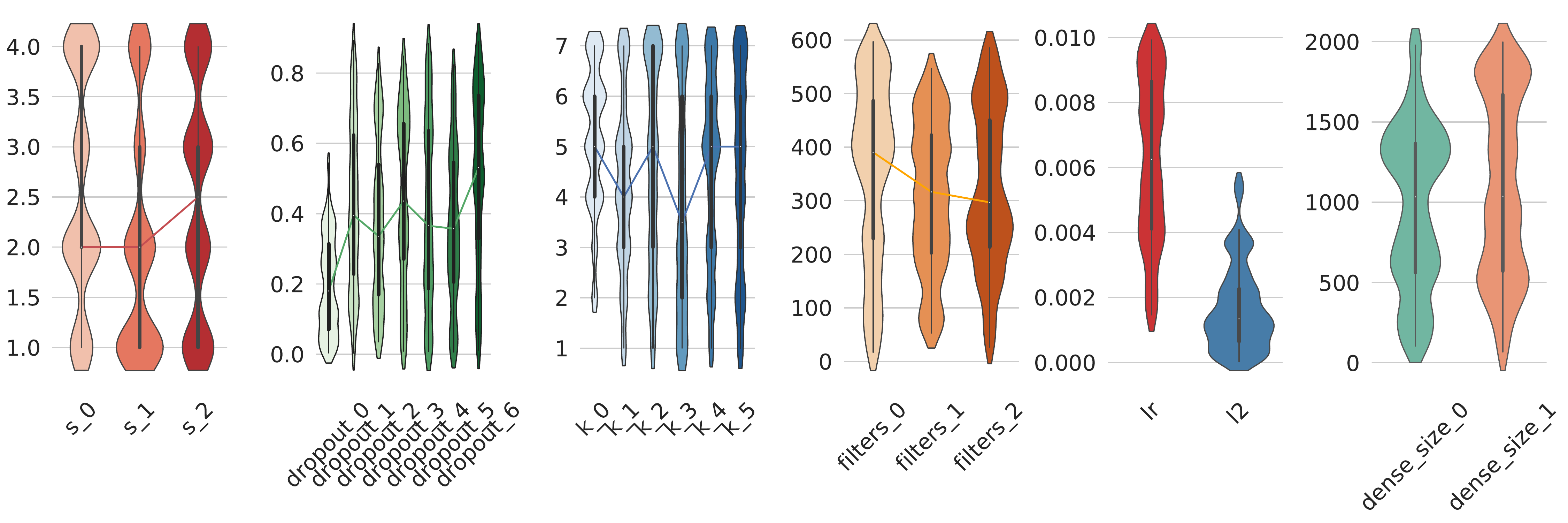}
\caption{On MNIST data set, the estimated distribution of hyper-parameters from the top-50 architectures. The distribution (violin) is obtained from kernel density estimation.}\label{fig:violin-mnist}
\end{figure*}

In Figures \ref{fig:MDS-mnist1}, \ref{fig:MDS-fashion1} and \ref{fig:MDS-cifar1}, \emph{Multi Dimensional Scaling} (MDS) \cite{kruskal1978multidimensional} is used to visualise the accuracy optimization landscape of the three problem instances. It can be observed, that CIFAR-10 is apparently a much harder problem than MNIST, since the mean accuracy of the landscape is much less. It can also be observed that the landscape visualisations are very similar in characteristics. This lead us to believe that the different NAS problems on image classification tasks can be solved using a general approach. This is, off course, also in line with transfer learning and neural architecture transfer. Using this design of experiments, we analyse the top 50 performing networks for each problem instance and look at the distribution of each search space parameter. In Figure \ref{fig:violin-mnist}, the parameters of the best 50 performing networks are visualised on the MNIST problem. The median setting is visualized by a line in each subplot.

\begin{figure}[!ht]
\centering
\setlength{\tabcolsep}{1pt}
\begin{tabular}{ll}
    \rotatebox{90}{\small MNIST} & \includegraphics[width=.7\textwidth,trim=0mm 25mm 0mm 0mm,clip]{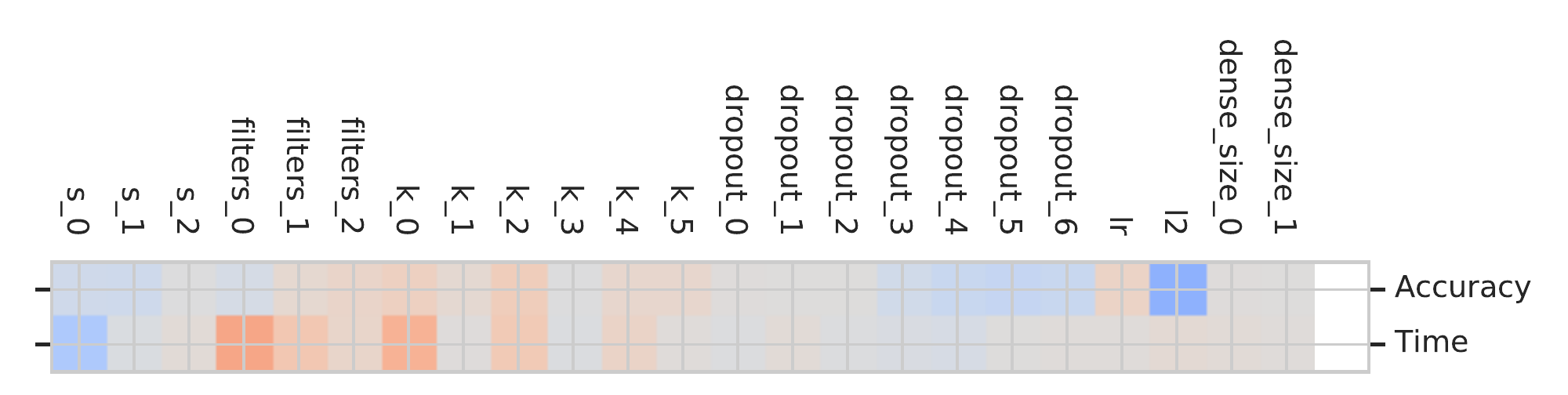} \\
    \rotatebox{90}{\small Fashion} &\includegraphics[width=.7\textwidth,trim=0mm 25mm 0mm 35mm,clip]{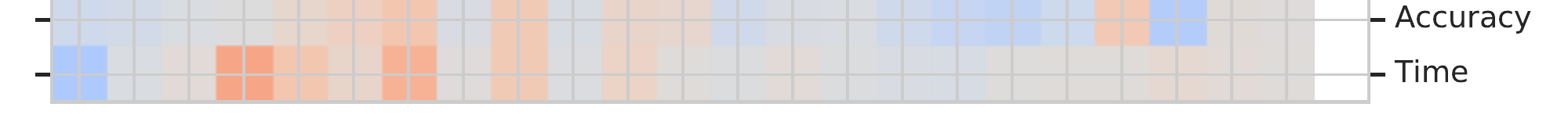} \\
    \rotatebox{90}{\hspace{10mm}\small CIFAR-10} & \includegraphics[width=.7\textwidth,trim=0mm 0mm 0mm 35mm,clip]{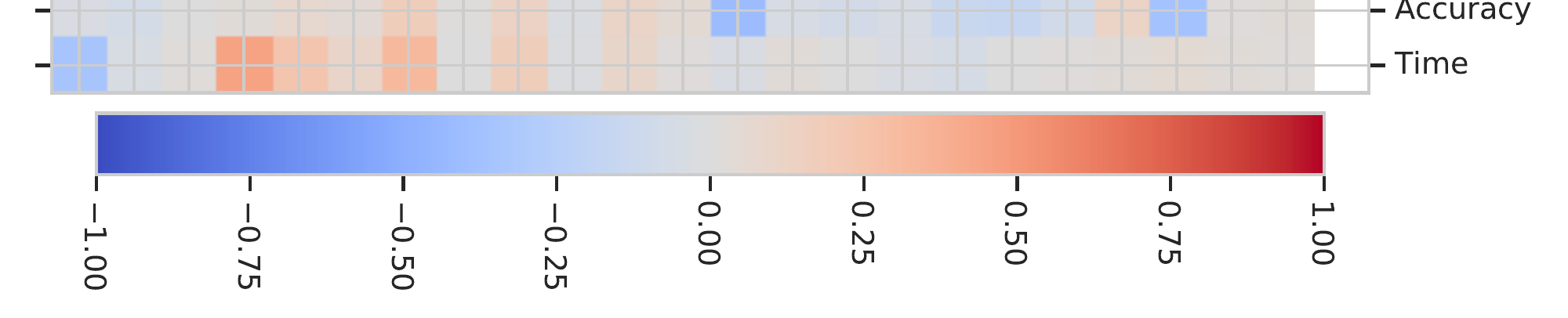} \\
\end{tabular}
\caption{Pearson's correlation coefficient between the accuracy and CPU time calculated on the architecture sampled in the initial range for data sets MNIST, Fashion, and CIFAR-10, respectively (top to bottom).\label{fig:corr}}
\end{figure}

To analyse which features are important for the performance of the neural architectures, the Pearson's correlation between each parameter and the CPU-time and accuracy are calculated and shown in Figure \ref{fig:corr}. The correlations of the features between each data set are almost identical. It is clear that the kernel size of the first layer and the dropout in the first layers are important features for the accuracy, and that there is a positive correlation between the number of filters and the CPU time (which is expected). 

Using the distribution from the top 50 network architectures (top 5\%), a new search space is defined as specified in Table~\ref{tab:searchspaces}. The new search space is a fraction of the original search space, and much more limited, especially for the dropout in the first layers and the learning rate.

Using this ``Reduced Range" search space, a new DOE of another $1000$ samples is generated and evaluated. 
In Figure \ref{fig:MDS-mnist2}, \ref{fig:MDS-fashion2} and \ref{fig:MDS-cifar2}, the MDS visualisations of the new optimization landscapes are shown. It can be observed that the new DOE has superior performance. If we take a look at the accuracy of all $1000$ samples from the MNIST neural architecture search, only two of the samples did not perform well, all others are near $100\%$ validation accuracy. We can observe that the architectures from this ``Reduced Range" search space are all having a very high accuracy among all three problem instances. Neural Architecture Search might not even be required in this case. The landscape seems to be very rugged but also robust, showing only small performance changes across a wide set of different architectures. As mentioned in \cite{schwartz2019green}, gaining a small accuracy increase by performing an expensive NAS procedure is in most cases not worth the time and effort. By starting with a well-performing base architecture, this effort can be significantly reduced or even omitted.
\begin{figure*}[!t]
\centering
\includegraphics[width=0.95\textwidth]{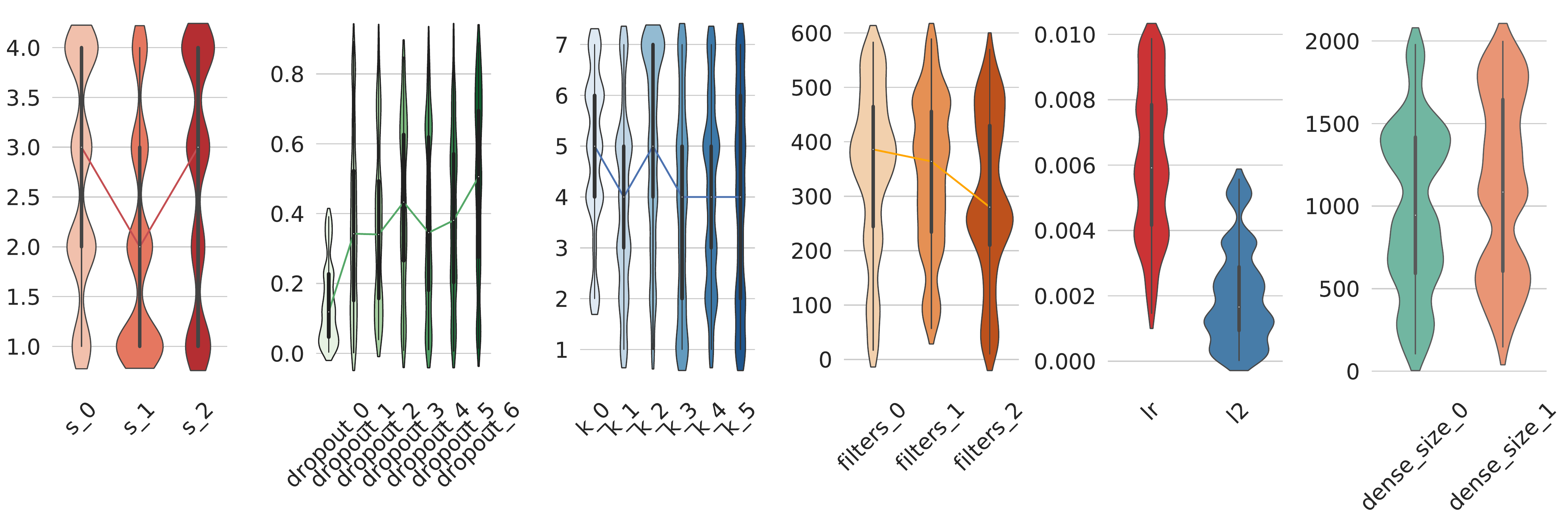}
\caption{On Fashion data set, the estimated distribution of hyper-parameters from the top-50 architectures. The distribution (violin) is obtained from kernel density estimation.}\label{fig:violin-fashion}
\end{figure*}
\begin{figure*}[!t]
\centering
\includegraphics[width=0.95\textwidth]{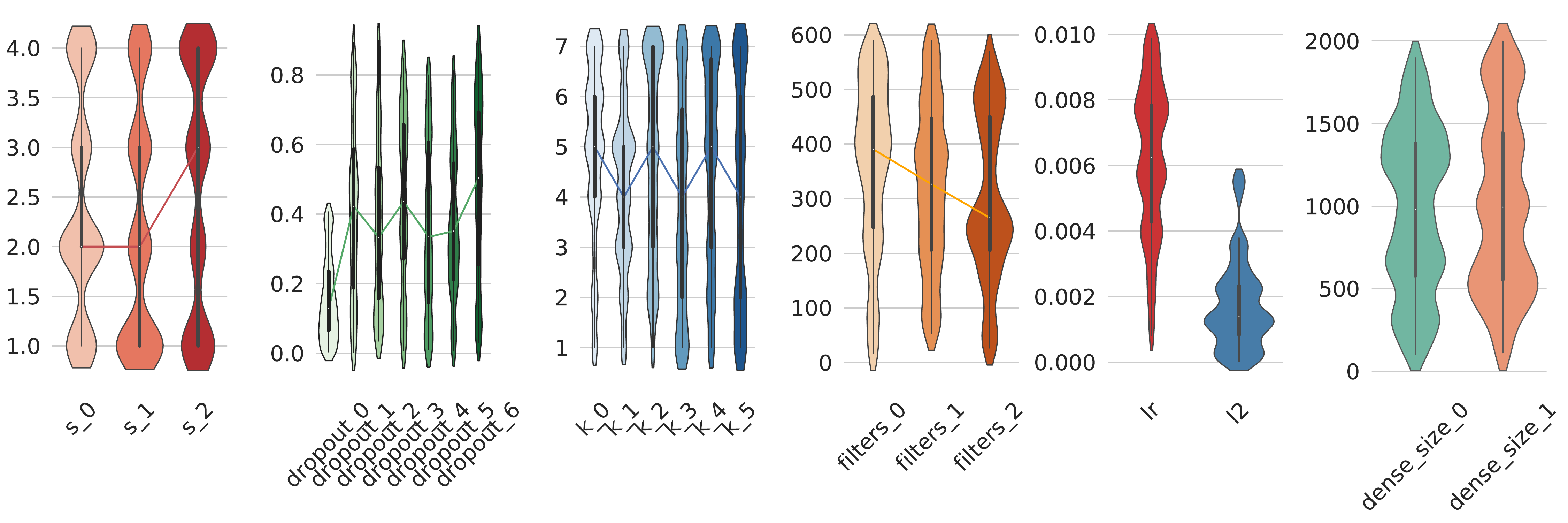}
\caption{On CIFAR-10 data set, the estimated distribution of hyper-parameters from the top-50 architectures. The distribution (violin) is obtained from kernel density estimation.}\label{fig:violin-cifar}
\end{figure*}

\section{Results on ELA features}
\begin{figure*}[!ht]
\centering
\includegraphics[width=1\textwidth]{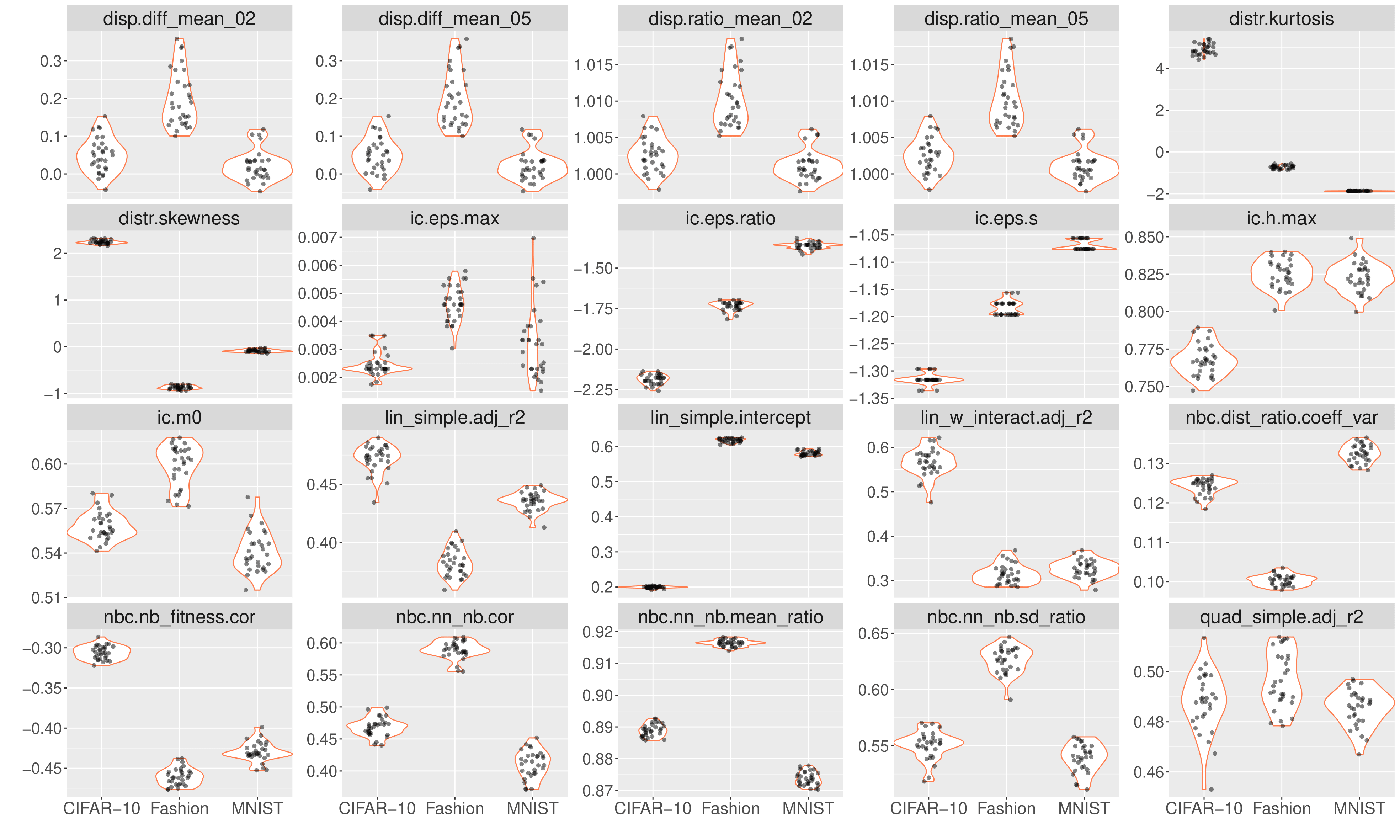}
\caption{On MNIST, CIFAR-10, and Fashion data sets, the distribution of the bootstrapped ELA features of the NAS landscape, which is computed on $1000$ architecture DoEs with the bootstrap size of $800$ and $30$ repetitions.}\label{fig:ela-bootstrap}
\end{figure*}

The aforementioned $20$ ELA features are computed on the $1000$ design points sampled from the ``Initial Range'' (the features obtained on the ``Reduced Range'' are very similar) (see Table~\ref{tab:searchspaces}). Also, to estimate the variability of feature values, we calculate those features with a bootstrapping procedure with the bootstrap size of $800$ and $30$ repetitions, the results of which are depicted as violin charts in Fig.~\ref{fig:ela-bootstrap}. Prior to the feature computation, we re-scale the hyper-parameters to the range $[-5,5]$ (this is to make the resulting feature comparable to those extracted on the BBOB problem set. See below). It is clear that most features entail a different statistical population on different training data sets, e.g., \texttt{distr.skewness} and \texttt{ic.eps.s}. For some features (e.g., \texttt{disp.ratio\_mean\_05}), although their absolute variation is marginal across different data sets, they contain enough power to distinguish the NAS landscape over the data sets. We found only \texttt{quad\_simple.adj\_r2} is not very informative according to this figure.

\begin{figure*}[!ht]
\centering
\subfloat[Bootstrapped ELA features of NAS landcapes]{\includegraphics[width=.5\textwidth,trim=0mm 25mm 0mm 10mm,clip]{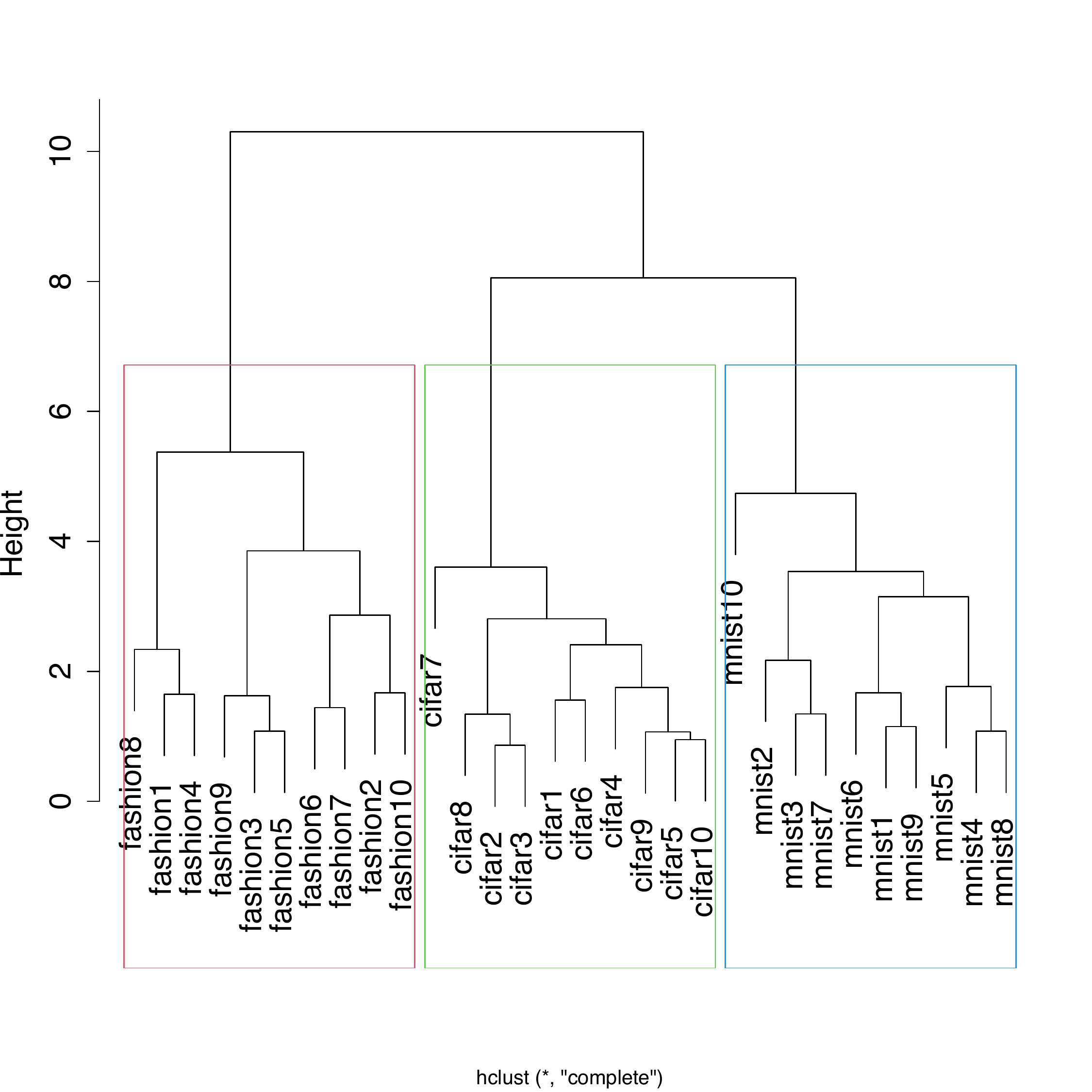}\label{fig:hclust-NN}}
\subfloat[ELA features of NAS + BBOB]{\includegraphics[width=.5\textwidth,trim=0mm 25mm 0mm 10mm,clip]{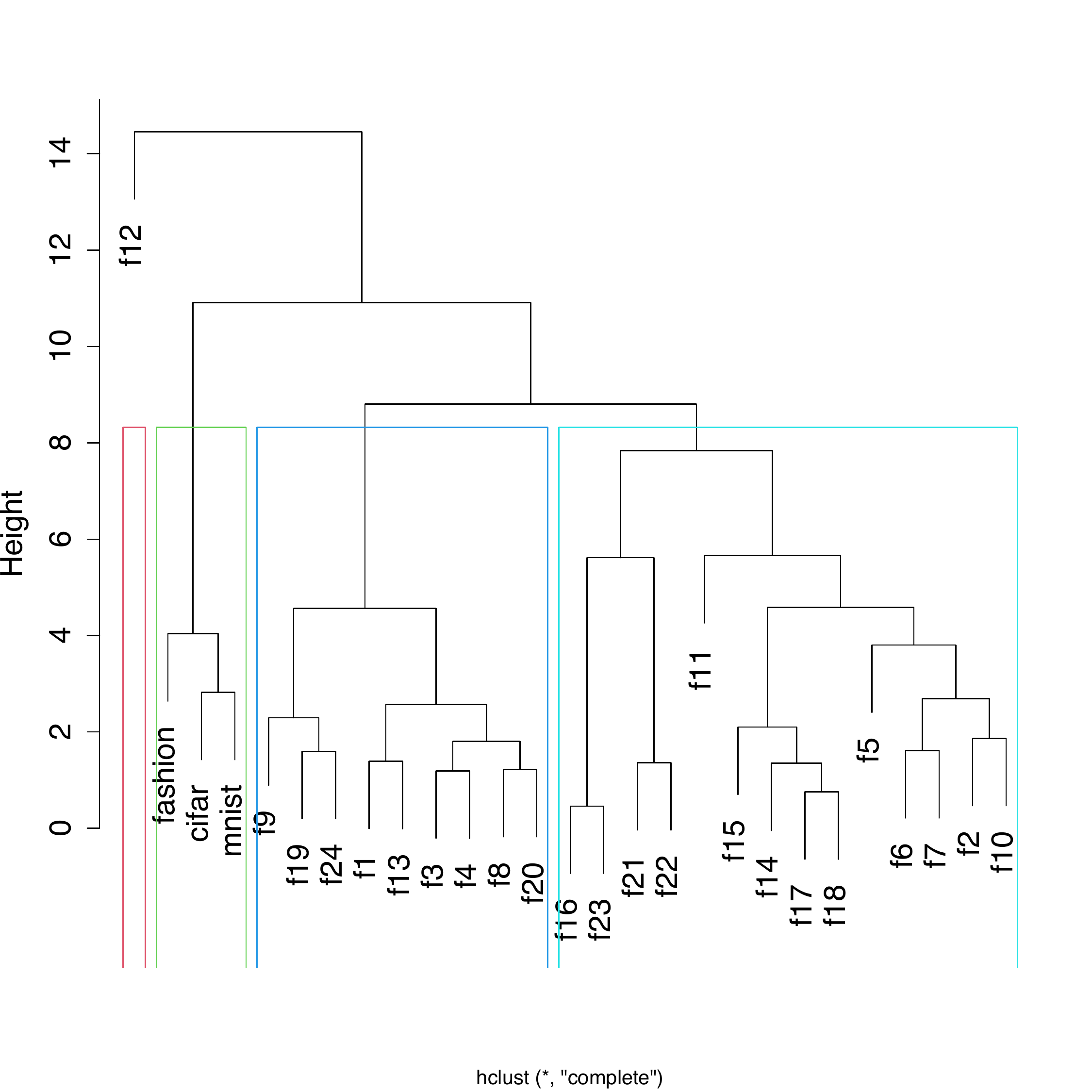}\label{fig:hclust-all}}
\caption{The clustering patterns given by the hierarchical clustering method with a proximity metric that takes the maximal Euclidean distance between two clusters. The ``Height'' on the $y$-axis indicates the proximity value for each pair of merged clusters. We additionally marked the major clusters using boxes.}
\end{figure*}

Based on the bootstrapped feature values, we subsequently delved into the clustering pattern that might exist when considering all three data sets together. As illustrated in Fig.~\ref{fig:hclust-NN}, we observed a crystal clear separation among the NAS landscapes on those three data sets, after performing a hierarchical clustering on all $30$ bootstrapped feature vectors (Note that, only $10$ feature vectors are rendered for each data set in order to make the plot more visually perceivable). In addition, the Euclidean distance is taken as the proximity metric between clusters and it is also obvious that the MNIST and CIFAR-10 data sets are more similar, compared to Fashion.

Naturally, we are also curious about whether the NAS landscape would resemble that of some conventional benchmark problem in terms of feature vectors, since if this is the case, then an optimizer that performs extremely well on such a benchmark problem will also be competitive on the NAS task. For this purpose, we choose the well-known \textbf{Black-Box Optimization Benchmarking} (BBOB)~\cite{elhara2019coco} problem set, which encompasses $24$ continuous problems of different kinds (e.g., (non-)separable, multi-modal, and highly rugged), and compute the same set of ELA features on all $24$ problems with $1000$ design points obtained from the LHS method. To make the results comparable, we set the dimensionality of BBOB problems to $23$ and apply the LHS method in the hyper-box $[-5,5]^{23}$ (since it is the range on which those problems are defined). The ELA features for our NAS instances are compared to 20 instances of all 24 functions from the BBOB benchmark~\cite{elhara2019coco}. In detail, we extracted the ELA features on the first $20$ 
problem instances (by which the original problem is randomly rotated and translated) of each BBOB problem, for improving the robustness of the resulting feature values.

We performed the same hierarchical clustering procedure again for both the NAS and BBOB landscape features, on the mean feature values over $30$ bootstraps (NAS) and $20$ problem instances (BBOB). The resulting clusters are shown in Fig.~\ref{fig:hclust-all}, in which we can see that the NAS landscape on three data sets falls into a category on its own and the other BBOB problem clusters are considerably distant to it (supported by the proximity metric on the $y$-axis).

When looking at all problem instances without any aggregation, we observed that the $20$ closest neighbours to MNIST have an Euclidean distance of $3.24 \pm 0.03$ while the average distance between the neighbours and their own $20$ neighbours is $0.71 \pm 0.17$. The similar pattern is also seen for CIFAR-10 ($3.24 \pm 0.04$ versus $0.72 \pm 0.18$) and Fashion ($2.71 \pm 0.04$ versus $0.72 \pm 0.20$). This seconds our previous findings from the hierarchical clustering that the NAS problems are not overlapping with any function group of the BBOB functions, with respect to the ELA features.

Among all BBOB problems, function $f12$, $f11$, and $f13$ are the nearest neighbours to all three of the NAS landscapes. On one hand, $f12$ is the \emph{Bent Cigar} function\footnote{See https://coco.gforge.inria.fr/}, which has a very narrow ridge that needs to be followed to optimize the function. On the other hand, $f11$ and $f13$ are classified as ``ill-conditioned" functions, suggesting that we would opt to apply, for the NAS task, optimizers devised particularly to handle the ill-conditioning scenario. According to the results of BBOB 2019\footnote{GECCO Workshop on Real-Parameter Black-Box Optimization Benchmarking}, this preference entails algorithms such as \emph{GLOBAL} (Sampling, clustering, and local search using BFGS or Nelder-Mead)~\cite{csendes1988nonlinear}, \emph{iAMALGAM} (Adapted Maximum-Likelihood Gaussian Model Iterated Density Estimation Algorithm 
)~\cite{bosman2013benchmarking}, and \emph{BIPOP-CMA-ES} (Bi-population Covariance Matrix Adaptation Evolution Strategy with random restarts)~\cite{hansen2009benchmarking}.

\section{Conclusions and outlook}
In this paper, we use the Exploratory Landscape Analysis (ELA) technique to probe into the landscape properties of the Neural Architecture Search (NAS) task. We designed a feedforward convolutional architecture with stacked structure for our purpose. In this way, the design/search space is narrowed down drastically for high accuracy solutions. Using the knowledge, of how the underlying optimization landscape behaves, better fitted algorithms can be used for NAS and a better starting point architecture can be developed from scratch. We have shown that for the three common image classification problems, MNIST, Fashion, and CIFAR-10, the NAS landscapes share many properties and the same basis neural architecture can be used to solve all three problem instances efficiently on a highly reduced search space without even requiring Neural Architecture Search.
We have reduced the ``Initial Range'' of the search space drastically to a fraction of the search space by comparing the distribution of the parameters of the best performing networks among all three NAS tasks. The network architectures resulting from the ``Reduced Range" search space showed supreme performance for all three image classification problems. \\
The ELA features can be used to compare NAS problems with continuous black-box optimization problems in order, for instance, to select a proper optimizer.
We showed that the ELA features of the NAS landscape are distinguishable across those three data sets, and are also perfectly separated from the ELA features of all $24$ BBOB problems, indicating that the NAS task is an entirely new problem category on its own when considering the dominant benchmark problem sets in the field of continuous black-box optimization. Hence, it would be beneficial, for algorithm selection and development, to study which optimizer of the state-of-the-art would perform well on the NAS task, and for the benchmark community, to augment the current problem sets for incorporating the NAS-like landscape.\\
In this paper the search space was limited to only real-valued and integer parameters, while categorical variables like the activation function were fixed because ELA features can only be calculated on continuous spaces. However, there are methods to also calculate features for categorical spaces that could be interesting to apply in a similar fashion.

\section*{Acknowledgments}
Our work was supported by the Paris Ile-de-France Region.

%
%
\bibliographystyle{unsrt}
\bibliography{main}
\end{document}